\documentclass[journal]{IEEEtran}
\ifCLASSINFOpdf
\else

\fi

\usepackage{times}  
\usepackage{helvet}  
\usepackage{courier}  
\usepackage{url}  
\usepackage{graphicx}  
\usepackage{amssymb}
\usepackage{amsmath}
\usepackage{multirow}
\usepackage{subfigure}
\begin{document}

\title{Reconstruction Regularized Deep Metric Learning for Multi-label Image Classification}

\author{Changsheng~Li,~\IEEEmembership{Member,~IEEE,}
        Chong~Liu,
        Lixin~Duan,
        Peng~Gao,
        Kai~Zheng,
\thanks{
Manuscript received October 31, 2018; revised March 21, 2019; accepted June 4, 2019. This work was supported in part by the NSFC under Grant 61806044, 61772118,  61836007, 61832017, 61532018 and in part by the Open Projects Program of National Laboratory of Pattern Recognition. (Corresponding author: C. Li; L. Duan.)

C. Li, C. Liu, L. Duan and K. Zheng are with School of Computer Science and Engineering and Big Data Research Center, University of Electronic Science and Technology of China. P. Gao is with PingAn Health Technology Co.,Ltd. E-mail: \{lichangsheng,zhengkai\}@uestc.edu.cn, \{andyliu281,lxduan\}@gmail.com, gaopeng712@pingan.com.cn.}
}

\markboth{IEEE Transactions on Neural Networks and Learning Systems}%
{Shell \MakeLowercase{\textit{et al.}}: Bare Demo of IEEEtran.cls for IEEE Journals}

\maketitle

\begin{abstract}
In this paper, we present a novel deep metric learning method to tackle the multi-label image classification problem.  In order to better learn the correlations among images features, as well as labels, we attempt to explore a latent space, where images and labels are embedded via two unique deep neural networks, respectively. To capture the relationships between image features and labels, we aim to learn a \emph{two-way} deep distance metric over the embedding space from two different views, i.e., the distance between one image and its labels is not only smaller than those distances between the image and its labels' nearest neighbors, but also smaller than the distances between the labels and other images corresponding to the labels' nearest neighbors.
Moreover, a reconstruction module for recovering correct labels is incorporated into the whole framework as a regularization term, such that the label embedding space is more representative.  Our model can be trained in an end-to-end manner.  Experimental results on publicly available image datasets corroborate the efficacy of our method compared with the state-of-the-arts.
\end{abstract}


\IEEEpeerreviewmaketitle

\section{Introduction}
In the past decade, multi-label learning has attracted lots of attention in the fields of neural network and machine learning \cite{charte2014li-mlc,liu2015on,liu2017an,Li2018A,shen2018multilabel}.
In this problem, instances (e.g., images, documents) are assumed to be associated with a set of labels instead of one single label.
In order to deal with this case, multi-label learning aims to learn a series of classifiers for labels, which can project an instance into a label vector with a fixed size.
Actually, multi-label learning is a special case of multi-output learning problems, where each label can be regarded as an output.
So far, multi-label learning has been widely applied to image classification \cite{Tan2015Learning}, text classification \cite{liu2017deep}, music instrument recognition \cite{xioufis2011multilabel}, and so on.
In this paper, we focus on solving the image classification problem by leveraging the multi-label learning technique.

To date, many multi-label learning approaches for image classification have been proposed \cite{Tan2015Learning,Li2014Multi,Luo2015Multiview,Wang2016CNN,Wei2016HCP,Zhao2016Regional,li2017improving,shen2018compact}.
Simply speaking, multi-label image classification can be achieved by casting this task into several binary-class subproblems, where each subproblem is to predict whether the image is relevant to the corresponding label. This kind of method takes different labels as independent ones. However, in practice, there are often correlations among labels, e.g., in an image of landscape, blue sky and white cloud often appear simultaneously. Empirically and theoretically speaking, taking advantage of such correlations during learning can help predict testing images more accurately \cite{Gao2013On,li2016self-paced,Wang2017Multi,li2019dynamic}.
Therefore, current mainstream approaches attempt to learn correlations among multiple labels based on training data and incorporate such correlations into the learning process for improving model performance \cite{Zhang2007ML}.  Here we briefly survey some typical algorithms (For a complete review, please refer to \cite{Zhang2014A}).
In \cite{Zhang2007ML}, a multi-label lazy learning approach, called Ml-knn, was presented, which utilized the maximum a posteriori (MAP) principle to determine label sets for unseen instances.
Authors in \cite{Luo2015Multiview} proposed a multi-view framework to fuse different kinds of features, and explored the complementary properties of different views for matrix completion based multi-label classification.
Huang and Zhou \cite{Huang2012Multi} proposed a method, i.e., multi-label learning using local correlation (MLLOC), which incorporated global discrimination fitting and local correlation sensitivity into a unified framework. Furthermore, Li et al. \cite{Li2018A} extended \cite{Huang2012Multi} into a self-paced framework, where the instances and labels were simultaneously learnt from an easy-to-hard fashion.

Recently, deep learning has achieved very promising results in various image applications, including object recognition/detection \cite{Ren2015Faster}, semantic segmentation \cite{Long2017Fully}, to name a few. In the meantime,  there are also some deep learning methods that are proposed for solving the multi-label image classification problems \cite{Tan2015Learning,Wang2016CNN,Li2016Correlated,Chen2017Recurrent,Yeh2017Learning,Wang2017Multi,he2018reinforced,liu2018multilabel,shen2018deep}.
For example, in \cite{Wang2016CNN}, authors proposed a CNN-RNN framework to jointly capture the semantic dependency among labels and the image-label relevance.
Authors in \cite{Zhu2017Learning} proposed a spatial regularization network to generate class-related attention maps and capture both spatial and semantic label dependencies.
In \cite{Yeh2017Learning}, authors integrated the deep canonical correlation analysis  and an autoencoder in a unified DNN architecture for image classification.
The method in \cite{Chen2017Recurrent} introduced a recurrent attention mechanism to locate attentional and contextual regions for multi-label prediction.

\begin{figure}
\centering
{\includegraphics[width=0.95\linewidth]{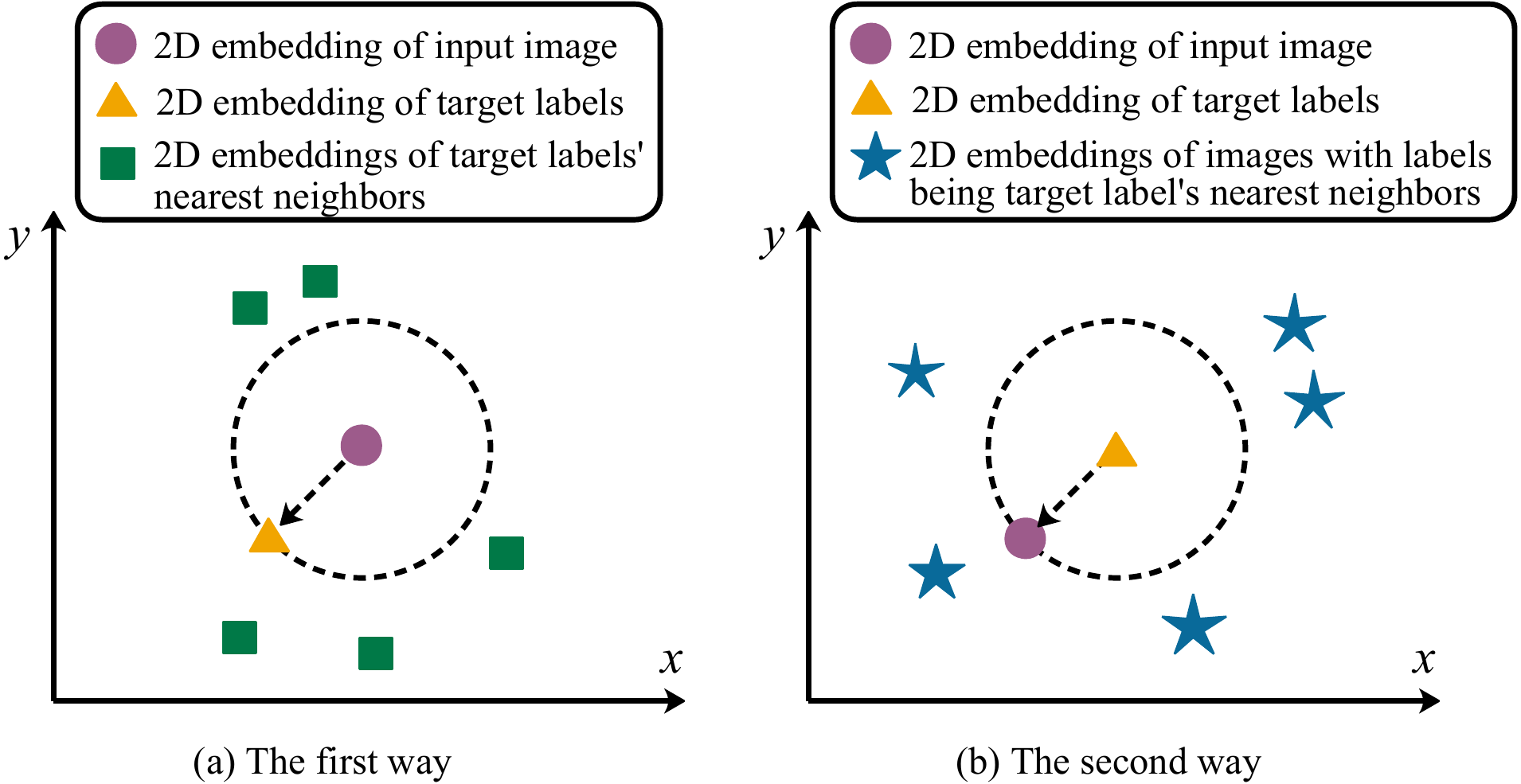}}
\caption{
 Schematic illustration of the main idea behind two-way distance metric learning. The x-y plane denotes the 2D embedding space.
}
\label{twoway}
\end{figure}

Different from the above methods, in this paper, we propose a novel framework for multi-label image classification, which is based on \underline{RE}construction regularized \underline{T}wo-way \underline{D}eep distance \underline{M}etric (RETDM) learning.
Specifically, we first attempt to learn an embedding space, where original images and labels are embedded via a Convolutional Neural Network (CNN) and a Deep Neural Network (DNN), respectively.
Through these two networks, we expect that image features dependency and labels dependency can be both discovered.
In order to capture the correlations between images and labels on the embedded space, a \emph{two-way} distance metric learning strategy is presented.
Figure \ref{twoway} illustrates the idea behind the two-way distance metric learning strategy.
In the embedding space, we hope the distance between an input image embedding vector and its label embedding vector is smaller than those distances between the image embedding vector and the embedding vectors of the labels' nearest neighbors as shown in Figure \ref{twoway}(a). In the meantime, as demonstrated in Figure \ref{twoway}(b), we also anticipate that the distance between the image embedding vector and its corresponding label embedding vector is smaller than those distances between the label embedding vector and other image embedding vectors with their labels being the nearest neighbors of the target labels.
By such way, two nearby instances with different labels will be pushed far away.
Finally, a reconstruction network is incorporated into the framework as a regularization term to make the learnt embedding space more representative.

Compared with state-of-the-art multi-label image classification methods, the proposed framework has the following advantages:
\begin{itemize}
\item An end-to-end trainable framework is proposed to integrate comprehensive distance metric learning into deep learning for multi-label image classification.
\item We present a two-way distance metric learning strategy based on two different views for capturing the correlations between images and labels, which is tailored for multi-label image classification.
\item A reconstruction error based loss function is introduced to regularize the label embedding space for further improving model performance.
\end{itemize}

We evaluate the proposed framework with exhaustive experiments on publicly available multi-label image datasets including scene, mirflickr, and NUS-WIDE. Experimental results demonstrate that the proposed method achieves significantly better performance compared to the state-of-the-art multi-label classification methods.

The remainder of this paper is organized as follows. Related works are reviewed in Section II, and the proposed RETDM is introduced in Section III. Extensive experiments are presented in Section IV, followed by conclusions in Section V.

\section{Related Work}
Our work is related to three lines of active research: 1) Shallow metric learning for multi-label prediction; 2) Deep metric learning for other image applications; 3) Deep learning for multi-label image classification.

\smallskip
\noindent\textbf{Shallow metric learning for multi-label prediction:}
\cite{Jin2010Learning} proposed a distance metric learning approach for multi-instance multi-label learning. The authors presented an iterative algorithm by alternating between the step of estimating instance-label association and the step of learning distance metrics from the estimated association.
In \cite{Zhang2012Maximum}, a maximum margin output coding (MMOC) formulation was proposed to learn a distance metric for capturing the correlations between inputs and outputs.
Although MMOC has shown promising results for multi-label prediction, it requires an expensive decoding procedure to recover multiple labels of each testing instance.
To avoid this issue, \cite{Liu2015Large,Liu2018Metric} incorporated $k$ nearest neighbor (kNN) constraints into a distance metric formulation, and provided a generalization error bound analysis to show that their method can converge to the optimal solution.
\cite{gouk2016learning} introduced linear and nonlinear distance metric learning methods, which aimed at improving the performance of kNN for multi-label data.
In \cite{verma2016a}, a novel metric learning framework was presented to integrate class-specific distance metrics and explicitly take into account inter-class correlations for multi-label prediction.
All the methods mentioned above aim to learn various shallow distance metric models for multi-label tasks. However, they do not incorporate deep learning, a very powerful tool for image analysis, into their framework.

\smallskip
\noindent\textbf{Deep metric learning for other applications:} Up to now, there have been many deep metric learning approaches proposed for various image tasks. For example, \cite{Chopra2005Learning} proposed a Siamese Network to learn complex similarity metrics for face verification.  The learning process minimized a discriminative loss function that drove the distance to be small for pairs of faces from the same individual, and large for pairs from different individuals.
Now the Siamese Network has been very popular for numerous applications beyond face verification \cite{bertinetto2016fully,Shen2017Deep}.
In \cite{Schroff2015FaceNet}, a triplet loss was introduced to directly learn an embedding into an Euclidean space for face verification, and two kinds of strategies for triplet selection were provided during training.
\cite{sohn2016improved} proposed a new metric learning objective called multi-class $N$-pair loss.
The proposed objective function generalized the triplet loss by allowing joint comparisons among more than one negative example, and reduced the computational burden of evaluating deep embedding vectors via an efficient batch construction strategy.
\cite{Song2016Deep} described a deep feature embedding and metric learning algorithm for image clustering/retrieve. The authors defined a novel structured prediction objective on the lifted pairwise distance matrix within the batch during the neural network training.
Soon afterwards, \cite{song2017deep} further proposed a novel framework for image clustering/retrieve which optimized the deep metric embedding with a learnable clustering function and a clustering metric in an end-to-end fashion.
Moreover, \cite{Movshovitzattias2017No} presented two proxy assignment schemes for optimizing the triplet loss on a different space of triplets, so that the computational cost of training DNN models can be reduced and the accuracy of the model can be improved.
In addition, in \cite{Wang2017Deep}, a novel angular loss was introduced based on the angle constraints of the triplet triangle instead of distance constraints.
Although these methods have shown encouraging results for various applications, they can not be directly applied to multi-label image classification. This is because they do not consider the correlations among labels at all.
\begin{figure*}
\centering
{\includegraphics[width=0.95\linewidth]{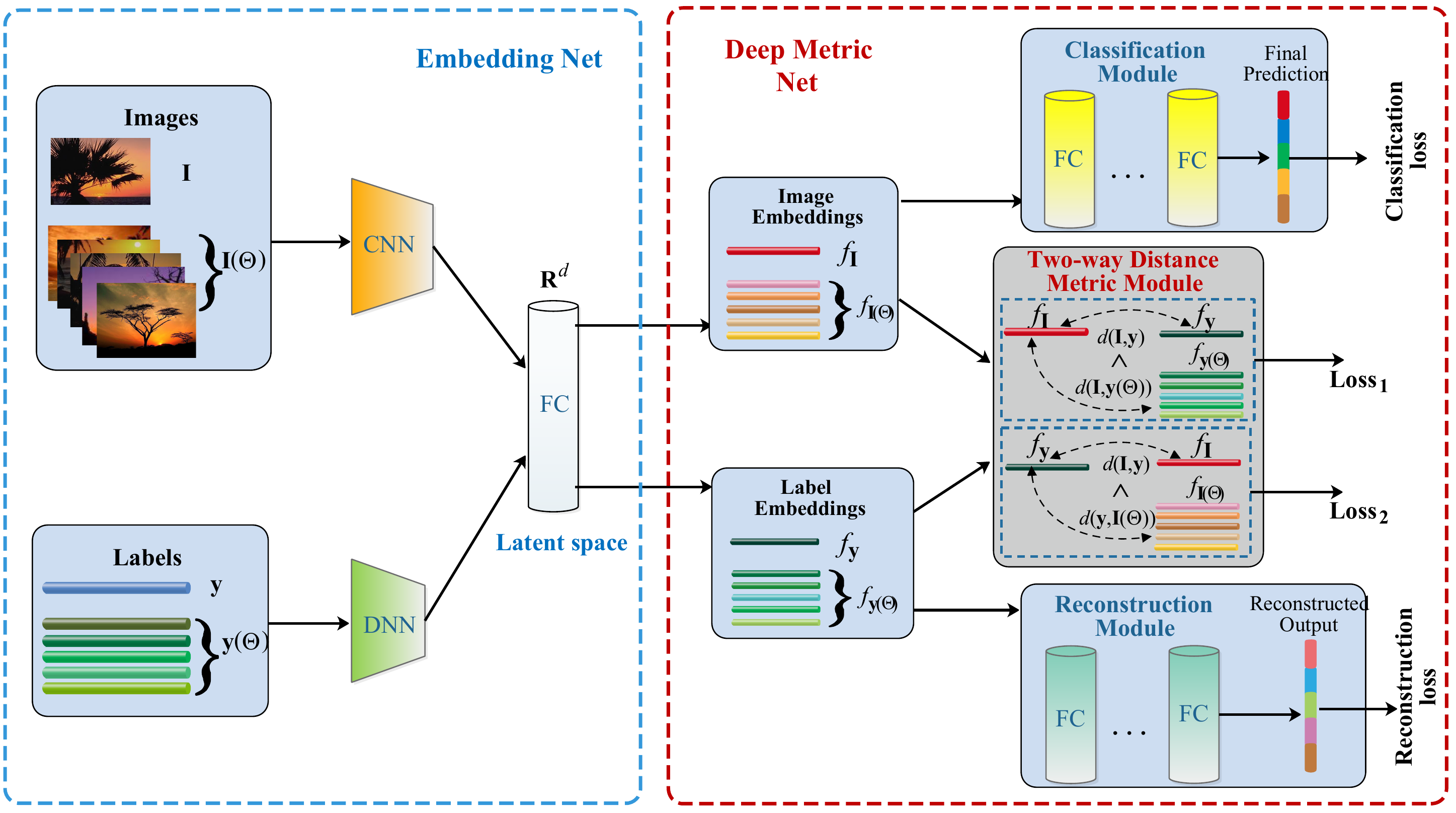}}
\caption{
Overall framework of our approach.
(Left) The embedding net (EN) consists of a convolutional neural network (CNN) and a deep neural network (DNN) to map images and labels to a latent space, respectively. VGG-16 is used as the CNN base model. DNN is comprised of two fully-connected layers. The dimension $d$ of the latent space is 512.
(Right) The deep metric net (DMN) contains three modules: one two-way distance metric module to learn the correlations between images and labels based on the latent space; one reconstruction module to regularize the label embedding space; one classification module used to make the image embedding space discriminative and predict unseen testing images. $\mathbf{y}$ is the target label of image $\mathbf{I}$;$\mathbf{y}(\Theta)$ denotes the set of $\mathbf{y}$'s $k$ nearest neighbors; $\mathbf{I}(\Theta)$ denotes the set of images corresponding to $\mathbf{y}$'s $k$ nearest neighbors.
}
\label{framework}
\end{figure*}

\smallskip
\noindent\textbf{Deep learning for multi-label image classification:}
Recently, deep learning has been gradually applied to multi-label image classification. For example,
\cite{Gong2014Deep} proposed to use ranking to train deep convolutional neural networks for multi-label image annotation problems.
In \cite{Tan2015Learning}, authors proposed a clique generating machine to learn graph structures, so as to exploiting label dependency for multi-label image classification.
The method in \cite{Wang2016CNN} formulated a CNN-RNN framework to jointly characterize the semantic dependency among labels and the image-label relevance.
\cite{Zhu2017Learning} further proposed a Spatial Regularization Network that generated class-related attention maps and captured both spatial and semantic label dependencies.
In \cite{Yeh2017Learning}, authors proposed Canonical Correlated Autoencoder (C2AE) for solving the task of multi-label classification. They integrated deep canonical correlation analysis  and autoencoder in a unified DNN model, and introduced sensitive loss functions to exploit cross-label dependency.
\cite{Chen2017Recurrent} introduced the recurrent attention mechanism into generic multi-label image classification for locating attentional and contextual regions regarding classification.
The authors in \cite{liu2018multilabel} boosted classification by distilling the unique knowledge from weakly-supervised detection into classification with only image-level annotations, and obtained promising results.

\section{Proposed Method}

We propose a deep neural network for multi-label classification, which takes advantage of distance metric learning to capture the dependency of image features, the dependency of labels, as well as the correlations between images and labels. The overall framework of our approach is shown in Figure \ref{framework}. Our framework consists of two main network structures: The embedding net (EN) and the deep metric net (DMN). The embedding net is used to embed images and labels into a latent space, and it consists of a convolutional neural network (CNN) and a deep neural network (DNN). The deep metric net (DMN) contains three modules: one two-way distance metric module to learn the correlations between images and labels over the latent space, which is tailored for multi-label image classification; one reconstruction module for regularizing the label embedding space;
one classification module used for making the image embedding space more discriminative and conducting predictions on unseen testing images.
The whole framework is trained in an end-to-end manner.

Let $\mathcal{S}=\{\mathbf{I}_i, \mathbf{y}_i\}_{i=1}^n$ denote a set of training data.  $\mathbf{I}_i$ is the $i$-th input image with ground-truth labels $\mathbf{y}_i = [y_i^1 , y_i^2 , ..., y_i^m]^T$, where $y_i^j$ is a binary indicator. $y_i^j = 1$ indicates image $\mathbf{I}_i$ is tagged with the $j$-th label, and $y_i^j = 0$ otherwise. $n$ and $m$ denote the number of all training images and all possible labels, respectively. The goal of multi-label image classification is to learn a serious of classifiers for mapping $\mathbf{I}_i$ to a vector $\mathbf{\widehat{y}}_i$, such that $\mathbf{\widehat{y}}_i$ is close to $\mathbf{y}_i$ as much as possible.

To achieve this goal, a simple model is to learn a projection matrix via minimizing the following logistic loss function:
\begin{align}\label{aa}
\min \mathcal{L}(\mathbf{I}_i,\mathbf{y}_i)= \sum_{i=1}^n \sum_{j=1}^m \log(1 + \exp({y}_i^j \mathbf{P} g(\mathbf{I}_i)))
\end{align}
where $g(\mathbf{I}_i)$ denotes the feature representation of image $\mathbf{I}_i$, where $g(\cdot)$ can be an arbitrarily hand-crafted feature extractor or deep learning based feature extractor. $\mathbf{P}$ is a learnable matrix to project $g(\mathbf{I}_i)$ to a new feature space. However, Eq. (\ref{aa}) treats all labels as independent ones, thus ignoring the dependency among labels.
In this paper, we utilize two deep neural networks to respectively embed images and labels to a latent space for discovering input dependency and labels dependency simultaneously. Based on the embedded space, a two-way distance metric is learned to capture the correlations between images and labels for multi-label image classification.

\smallskip
\noindent
\textbf{Learning an embedding space for images and labels}

\noindent For obtaining an input image embedding vector, image $\mathbf{I}_i$ is first resized to $W\times H$ and fed into a convolutional neural network. In this paper, we use VGG-16 \cite{Simonyan14c} as the base model and $W\times H$ is thus set to $224\times 224$. We dropped the last fully-connected layer in the base model, and changed the number of hidden units of the second fully-connected layer to 512. Then the image embedding vector $f_{\mathbf{I}_i}\in \mathbf{R}^d$ can be represented by:
\begin{align}
f_{\mathbf{I}_i} = \Phi_{cnn} (\mathbf{I}_i, \theta_{cnn})
\end{align}
where $\Phi_{cnn}$ is the architecture of the image embedding network, and $\theta_{cnn}$ is the learnt parameters of the network.

In order to acquire the label embedding vector, $\mathbf{y}_i$ is fed into a deep neural network which consists of two fully-connected layers with 512 hidden units. The label embedding vector can be represented by
\begin{align}
f_{\mathbf{y}_i} = \Phi_{dnn} (\mathbf{y}_i, \theta_{dnn})
\end{align}
where $f_{\mathbf{y}_i} \in \mathbf{R}^d$ has the same dimension with $f_{\mathbf{I}_i}$. $\Phi_{dnn}$ is the architecture of the label embedding network, and $\theta_{dnn}$ is learnable parameters of the network.

Through the networks $\Phi_{cnn}$ and $\Phi_{dnn}$, we expect to learn an embedding space to discover dependencies from image features and labels, respectively.
In the meantime, it is desired that the correlations between images and labels can be captured on this latent space. To reach this goal, we integrate three modules operated on the latent space into the whole framework: one two-way distance metric module which is specially designed for multi-label image classification, one reconstruction module aiming at regularizing the embedding space, and one classification module used for predicting labels during the inference phase.

\smallskip
\noindent
\textbf{Two-way Distance Metric Module}

\noindent
 Based on the embedding space, it is expected that the correlations between images and labels can be captured, and can be integrated into the learning process, such that the distance between an image embedding vector and its corresponding label embedding vector is not only smaller than the distances between the image embedding vector and the embedding vectors of the target labels' nearest neighbors, but also smaller than the distances between the label embedding vector and other images with labels being the target labels' nearest neighbors.

\begin{figure}
\centering
{\includegraphics[width=0.997\linewidth]{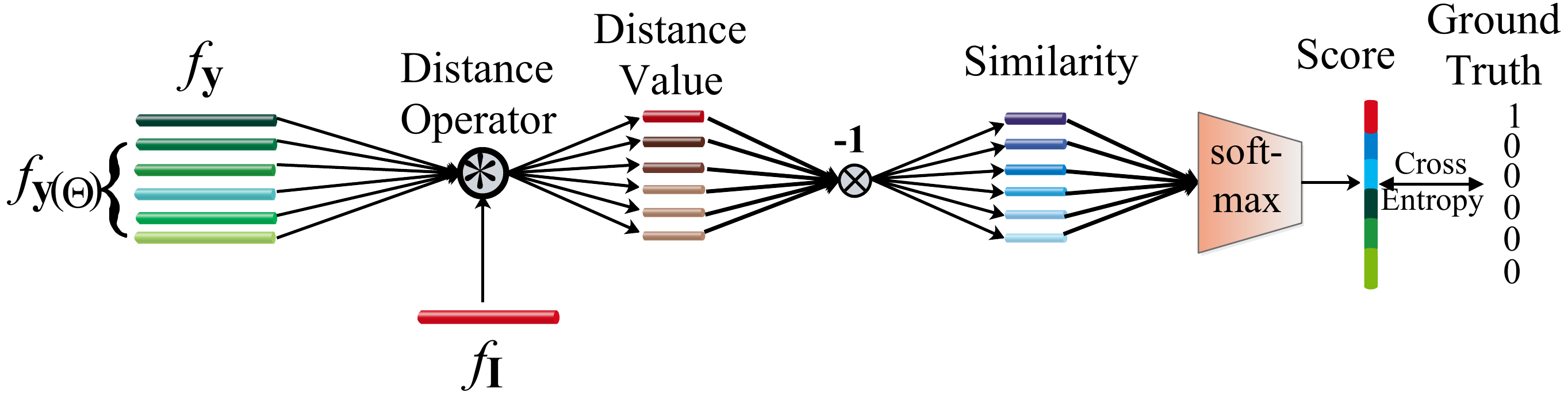}}
\caption{
An illustration for one-way deep distance metric learning.
}
\label{oneway}
\end{figure}

In sight of these, we propose a two-way strategy for deep distance metric learning. We first give a detailed description for one-way deep distance metric learning, as shown in Figure \ref{oneway}. Let ${\mathbf{y}}$ be the label of image ${\mathbf{I}}$. $f_{\mathbf{I}}$ denotes the embedding vector of image $\mathbf{I}$; $f_{\mathbf{y}}$ and $f_{\mathbf{y(\Theta)}}$ denote the embedding vectors of label $\mathbf{y}$ and its $k$ nearest neighbors $\mathbf{y(\Theta)}$, respectively. In order to make the distance between $f_{\mathbf{I}}$ and $f_{\mathbf{y}}$ be smaller than the distance between $f_{\mathbf{I}}$  and $f_{\mathbf{y(\Theta)}}$, the following constraints should be satisfied:
\begin{align}
d(f_{\mathbf{I}},f_{\mathbf{y}})<d(f_{\mathbf{I}},f_{\mathbf{y}_i}), \forall \mathbf{y}_i\in \mathbf{y}(\Theta)\label{constraint}
\end{align}
where $d(\cdot)$ is an arbitrary distance function. Here the Euclidean distance is chosen in the experiment.

In order to satisfy the constraints in (\ref{constraint}), we formulate it as a multi-class classification problem. The vector $[(f_{\mathbf{I}},f_{\mathbf{y}}), (f_{\mathbf{I}},f_{\mathbf{y}^1}),\ldots, (f_{\mathbf{I}},f_{\mathbf{y}^k})]$ is regarded as a sample, where $\mathbf{y}^i\in \mathbf{y}(\Theta)$ and $\cup_{i=1}^k \mathbf{y}^i =\mathbf{y}(\Theta)$. The new sample's label vector is [1, 0, \ldots, 0].  The distance values are taken as the feature representation of the new sample.

After obtaining the distances between image embedding vector $f_{\mathbf{I}}$ and label embedding vectors $f_{\mathbf{y}}$ and $f_{\mathbf{y}(\Theta)}$. We then can calculate the similarity scores between image and labels by
\begin{align}
sim(\mathbf{I},{\mathbf{y}^i})=-1* d(f_{\mathbf{I}},f_{{\mathbf{y}^i}}), \forall {\mathbf{y}^i}\in \{\mathbf{y} \cup \mathbf{y}(\Theta)\}
\end{align}

Our goal is to maximize the score $sim(\mathbf{I},{\mathbf{y}})$, while minimize the scores $sim(\mathbf{I},{\mathbf{y}(\Theta)})$, so that the distance between image $\mathbf{I}$ and its target output $\mathbf{y}$ is the smallest in the embedding space.
Thus, given these similarity scores, our network produces a distribution for image $\mathbf{I}$ based on a softmax over these scores in the embedding space:
\begin{align}
p(z=1|f_{\mathbf{I}},f_{\mathbf{y}},f_{\mathbf{y}(\Theta)})=\frac{\exp (sim(\mathbf{I},\mathbf{y}))}{\sum_{{\mathbf{y}^i}\in \{\mathbf{y}\cup\mathbf{y}(\Theta)\}} \exp (sim(\mathbf{I},{\mathbf{y}^i }))} \nonumber
\end{align}

Finally, learning can be proceeded by minimizing the negative log-probability as
\begin{align}
\mathcal{L}_{1}= - \text{log}\  p(z=1|f_{\mathbf{I}},f_{\mathbf{y}},f_{\mathbf{y}(\Theta)}) \nonumber
\end{align}

In order to penalize the case that $d(f_{\mathbf{I}},f_{\mathbf{y}})$ is greater than or equal to $d(f_{\mathbf{I}},f_{\mathbf{y}_i})$, we propose a new loss function to be minimized as
\begin{align}\label{loss11}
\mathcal{J}_{1}=\left\{
\begin{aligned}
&\mathcal{L}_{1}, \ \ \text{if} \ d(f_{\mathbf{I}},f_{\mathbf{y}})\geq d(f_{\mathbf{I}},f_{\mathbf{y}_i}), \exists \mathbf{y}_i\in \mathbf{y}(\Theta)\\
&0,\ \ \text{otherwise}
\end{aligned}
\right.
\end{align}

For the second way distance metric learning, we expect that the distance between $f_{\mathbf{y}}$ and  $f_{\mathbf{I}}$ is smaller than the distances of  $f_{\mathbf{y}}$ and $f_{\mathbf{I}(\Theta)}$, where $\mathbf{I}(\Theta)$ denotes the set of images with labels being $\mathbf{y}$'s $k$ nearest neighbors.
The whole process is similar to that in the first way distance metric learning. Therefore, we can obtain another probability distribution over new similarity scores:
\begin{align}
p(z=1|f_{\mathbf{I}},f_{\mathbf{y}},f_{\mathbf{I}(\Theta)})=\frac{\exp (sim(\mathbf{y},\mathbf{I}))}{\sum_{{\mathbf{I}^i}\in \{\mathbf{I}\cup\mathbf{I}(\Theta)\}} \exp (sim(\mathbf{y},{\mathbf{I}^i}))} \nonumber
\end{align}
where $sim(\mathbf{y},\mathbf{I})= d (f_{\mathbf{y}}, f_{\mathbf{I}})$, and $sim(\mathbf{y},{\mathbf{I}^i})=d(f_{\mathbf{y}},f_{\mathbf{I}_i})$.

Thus, we minimize another loss function  as
\begin{align}\label{loss12}
\mathcal{J}_{2}=\left\{
\begin{aligned}
&\mathcal{L}_{2}, \ \ \text{if} \ d(f_{\mathbf{I}},f_{\mathbf{y}})\geq d(f_{\mathbf{I}_i},f_{\mathbf{y}}), \exists \mathbf{I}_i\in \mathbf{I}(\Theta)\\
&0,\ \ \text{otherwise}
\end{aligned}
\right.
\end{align}
where
\begin{align}
\mathcal{L}_{2}= - \text{log}\  p(z=1|f_{\mathbf{I}},f_{\mathbf{y}}, f_{\mathbf{I}(\Theta)}) \nonumber
\end{align}

Based on Eq. (\ref{loss11}) and Eq. (\ref{loss12}), we finally obtain a joint loss function for metric learning as
\begin{align}\label{loss1}
\mathcal{J}_{metric}= \mathcal{J}_{1} + \lambda \mathcal{J}_{2}
\end{align}
where $\lambda\geq 0$ is a hyper-parameters. In the experiment, we simply set $\lambda = 1$.

Through two-way distance metric learning, the latent space is more informative, and the correlations between images and labels can be well learned, which is beneficial for multi-label image classification.

\smallskip
\noindent
\textbf{Classification Module and Reconstruction Module}

\noindent In the field of spatio-temporal data mining, combining reconstruction loss with classification/regression tasks has been touched upon in recent studies \cite{Correlated,kieu2018distinguishing,kieu2018outlier}. Motivated by this, we jointly optimize the reconstruction loss and classification loss for multi-label image classification.

 In order to make the embedding space of images more discriminative, a classification module is introduced into the framework which conducts binary classification for each of the $m$ labels. It consists of one fully connected layer, followed by a sigmoid layer for each category.
\begin{align}
\widehat{\mathbf{y}} = \Phi_{cls}(f_{\mathbf{I}}, \theta_{cls}), \widehat{\mathbf{y}}\in \mathbf{R}^m
\end{align}
where $\theta_{cls}$ is the learned parameters of the classification module, and $\widehat{\mathbf{y}}=[\widehat{{y}}_1, \ldots, \widehat{{y}}_m]^T$ is predicted label confidences for each category. Prediction errors are measured via binary cross entropy over $\widehat{\mathbf{y}}$ and ground truth $\mathbf{y}$ as:
\begin{align}\label{loss2}
{\mathcal{J}_{cls}} = -\sum_{i=1}^m (y_i \text{log} (\widehat{{y}}_i) + (1-y_i) \text{log} (1-\widehat{{y}}_i))
\end{align}

Based on the label embedding space, a reconstruction module is incorporated into the whole architecture, so as to make the embedding more representative. The reconstruction module contains one fully connected layer for recovering $\mathbf{y}$. The reconstructed output $\bar{\mathbf{y}}$ can be expressed as:
\begin{align}
\bar{\mathbf{y}} = \Phi_{rec}(f_{\mathbf{y}}, \theta_{rec}), \bar{\mathbf{y}}\in \mathbf{R}^m
\end{align}
where $\theta_{rec}$ is the parameters of the reconstruction module, and $\bar{\mathbf{y}}=[\bar{{y}}_1, \ldots, \bar{{y}}_m]^T$ is the reconstructed output.
 We measure the reconstruction error through the mean square error (MSE) with respect to $\bar{\mathbf{y}}$ and ground truth $\mathbf{y}$ as:
\begin{align}\label{loss3}
{\mathcal{J}_{rec}} = \frac{1}{m}\sum_{i=1}^m (y_i -\bar{y}_i)^2
\end{align}

\smallskip
\noindent
\textbf{Overall Network and Training Scheme}

\noindent Based on these three modules, we propose to jointly minimize the following loss function:
\begin{align}
\mathcal{L}= \mathcal{L}_{cls}+\alpha \mathcal{L}_{metric} +\beta \mathcal{L}_{rec}
\end{align}
where $\alpha$ and $\beta$ are two hyper-parameters to balance the three losses.

The network can be trained by the following steps. First, we fine-tune only the classification net on the target dataset, i.e. setting $\alpha=\beta=0$, which is pre-trained on 1000-classification task of ImageNet dataset \cite{Deng2009ImageNet}.
Both $\Phi_{cnn}(\mathbf{I}_i,\theta_{cnn})$ and $\Phi_{cls}(f_{\mathbf{I}},\theta_{cls})$ are learned with cross-entropy loss.
Secondly, we fix $\Phi_{cnn}(\mathbf{I}_i,\theta_{cnn})$ and $\Phi_{cls}(f_{\mathbf{I}},\theta_{cls})$ , and focus on training $\Phi_{dnn}(\mathbf{y}_i,\theta_{dnn})$ and $\Phi_{rec}(f_{\mathbf{y}},\theta_{rec})$ with loss $\alpha \mathcal{L}_{metric} +\beta \mathcal{L}_{rec}$.
Finally, the whole network is jointly fine-tuned with loss $\mathcal{L}_{cls}+\alpha \mathcal{L}_{metric} +\beta \mathcal{L}_{rec}$.

Our deep neural network is implemented with Pytorch library\footnote{https://pytorch.org/}. In the experiment, we adopt the image augmentation strategies as suggested in \cite{Wang2015Towards}, which is a powerful tool to reduce the risk of over-fitting.
The input images are first resized to $256\times 256$, and then cropped at four corners and the center.
Finally, the cropped images are further cropped to $224\times 224$.
We employ stochastic gradient descend algorithm for training, with a batch size of 256, a momentum of 0.9, and weight decay of 0.0001.
The initial learning rate is set as 0.1, and decreased to 1/10 of the previous value whenever validation loss gets saturated, until 0.01 or the maximum epoch is reached.
We train our model with 8 NVIDIA Titan XP GPUs.
For testing, we simply resize all images to $224\times 224$ and conduct single-crop evaluation.

\section{Experiment}
In this section, in order to sufficiently verify the effectiveness of our method, RETDM, we perform it on three publicly available image datasets, scene \cite{Maron1998Multiple}, mirflickr \cite{huiskes08}, and Microsoft COCO \cite{Lin2014Microsoft}.
Table \ref{detail} lists the details of the three datasets.
These datasets are widely used for evaluating multi-label image classification algorithms.
Experimental results demonstrate that our proposed RETDM significantly outperforms the state-of-the-art methods on all the three datasets, and has strong generalization capability to different types of labels.

\begin{table}
\caption{Details of the used datasets.}
\centering
\begin{tabular}{|c|c|c|}
\hline
 {Dataset}  & Number of Images & Number of Labels \\
\hline
{scene}  &    2,000 &  5  \\
\hline
{mirflickr}  &    25,000 &  24  \\
\hline
{MS-COCO}  &    123,287 &  80  \\
\hline
\end{tabular}
\label{detail}
\end{table}

\begin{figure}
\centering
{\includegraphics[width=0.95\linewidth]{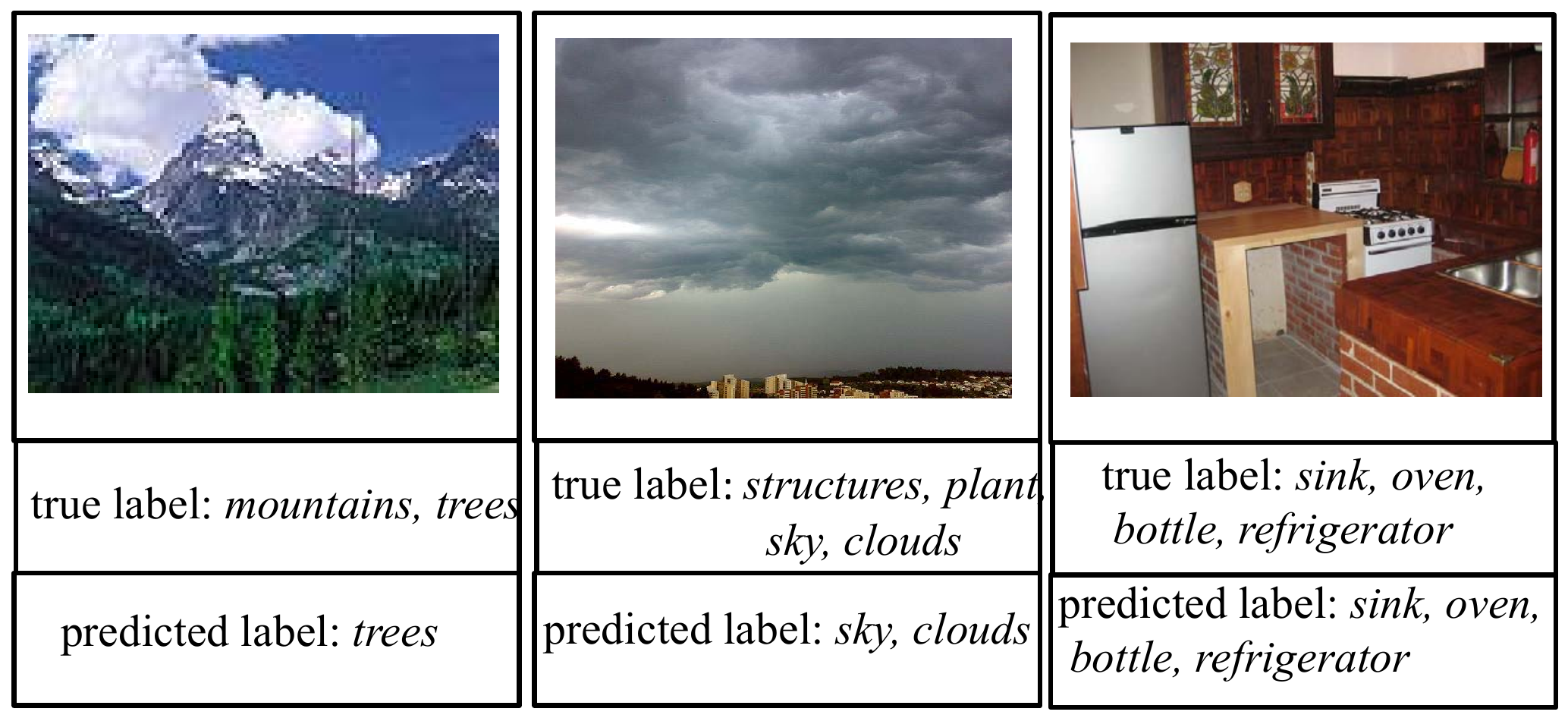}}
\caption{One example image from scene (left), mirflickr (middle) and MS-COCO(right) datasets, the ground-truth annotations and our
model's predictions.
}
\label{predict_true}
\end{figure}
\subsection{Experimental Datasets}

The scene image dataset consists of 2,000 natural scene images, where a set of labels is annotated to each image manually. There are five possible class labels, including desert, mountains, sea, sunset and trees.
On average, each image is associated with 1.24 class labels. An example of the annotations and predictions
for both of label set is shown in the left side of Fig. \ref{predict_true}.

The mirflickr image dataset \cite{huiskes08} contains 25,000 images that are representative of a generic domain and are of high quality. There are 24 possible labels in total, for instance ``sky'', ``water'', ``sea'', ``clouds'', and so on.
 The average number of labels per image is 8.94. In the dataset there are 1386 tags which occur in at least 20 images.
 An example of the annotations and predictions
for both of label set is shown in the middle of Fig. \ref{predict_true}.

 The Microsoft COCO (MS-COCO) dataset \cite{Lin2014Microsoft} is an image recognition, segmentation, and captioning dataset. Following \cite{Wang2016CNN}, we use it to evaluate multi-label learning algorithms.
The training set is composed of 82,783 images, which contain common objects in the scenes.
There are 80 classes with about 2.9 labels per image.
Another 40,504 images are employed as testing data in the experiment.
The number of labels for each image varies considerably on this dataset.
An example of the annotations and predictions
for both of label set is shown in the right side of Fig. \ref{predict_true}.

For the scene and mirflickr datasets, 80\% images are randomly chosen as the training data, and the rest are used for testing data. On the MS-COCO dataset, we use the same split of training/testing as \cite{Wang2016CNN} for a fair comparison.
\begin{table*}
\caption{Quantitative results by our proposed RETDM and compared methods on the scene dataset.}
\label{performance}
\centering
\begin{tabular}{c|c|c|c|c|c|c||c|c|c|c|c}
\hline
 {Method}  & C-P  & C-R  & C-F1  & O-P  & O-R  & O-F1 & hamming loss & ranking loss & coverage &  one error& average precision \\
\hline
{LMMO-kNN}  &     0.453 &  0.725   &   0.576    &  0.412    &  0.705 &    0.613  & 0.3985 & 0.6867 & 2.1450 & 0.7075 &0.5060 \\
{MLSPL}  &     0.425 &  0.682   &   0.563    &  0.402    &  0.699 &    0.610  & 0.4155 & 0.7214 & 2.2687 & 0.7089 &0.4561 \\
{CL}  &     \textbf{0.973} &  0.136   &   0.152    &  0.867    &  0.142 &    0.242  & 0.8603 & 0.2225 & 1.7225 & \textbf{0.0275} &0.6162 \\
BCE    &     0.906 & 0.861 & 0.883 & \textbf{0.906 }& 0.869  & 0.883 & 0.1331 & \textbf{0.0580} & 0.6200 &0.0475& 0.9223 \\
C2AE &    0.380 &  0.710  & 0.414  & 0.343 &  0.725 &  0.465 & 0.5552 & 0.8235 & 1.6675& 0.2625& 0.6241 \\
\hline
RETDM &   {0.893}  & \textbf{0.879} & \textbf{0.885} & 0.892  & \textbf{0.877}  & \textbf{0.884} & \textbf{0.1217 }& \textbf{0.0580}&  \textbf{0.5850}& 0.0425& \textbf{0.9285}\\
\hline
\end{tabular}
\label{scene}
\end{table*}

 \begin{table*}
\caption{Quantitative results by our proposed RETDM and compared methods on the mirflickr dataset.}
\label{performance}
\centering
\begin{tabular}{c|c|c|c|c|c|c||c|c|c|c|c}
\hline
 {Method}  & C-P  & C-R  & C-F1  & O-P  & O-R  & O-F1 & hamming loss & ranking loss & one error & coverage & average precision \\
 \hline
{LMMO-kNN}  &     0.433 &  0.602   &   0.463    &  0.577    &  0.615 &    0.601  & 0.9013 & 0.4025 & 0.0677 & 13.1261 &0.5664 \\
{MLSPL}  &     0.458 &  0.652   &   0.481    &  0.592    &  0.628 &    0.619  & 0.8235 & 0.3417 & 0.0518 & 13.0482 &0.5822 \\
{CL}  &     0.995 &  0.012   &   0.018    &  0.872    &  0.032 &    0.062  & 0.9626 & 0.1498 & 20.2414 & 0.0167 &0.1964 \\
BCE    &     0.735 &0.677& 0.704& 0.782& 0.739& 0.760& 0.2758& 0.0717 & 0.0344& 11.0742& 0.7076\\
C2AE &    0.476 &0.616 & 0.505 &0.613 & 0.630 & 0.621 & 0.9029& 0.3869 & 0.0660 &12.9678& 0.5767 \\
\hline
RETDM &   0.756 & 0.661 & 0.705 & 0.811 & 0.749 & 0.778 & 0.2645 & 0.0659 & 0.0296 & 10.8320 & 0.7275 \\
\hline
\end{tabular}
\label{mirflickr}
\end{table*}

 \begin{table*}
\caption{Quantitative results by our proposed RETDM and compared methods on the MS-COCO dataset.}
\label{performance}
\centering
\begin{tabular}{c|c|c|c|c|c|c||c|c|c|c|c}
\hline
 {Method}  & C-P  & C-R  & C-F1  & O-P  & O-R  & O-F1 & hamming loss & ranking loss &   coverage& one error& average precision \\
\hline
{LMMO-kNN}  &     0.298 &  0.379   &   0.342    &  0.401    &  0.438 &    0.410  & 0.5123 & 0.5582 & 37.4352 & 0.1734 &0.3877 \\
{MLSPL}  &     0.289 &  0.367   &   0.332    &  0.387    &  0.403 &    0.399  & 0.5334 & 0.5584 & 37.5626 & 0.1866 &0.3781 \\
{CNN-Softmax}  &     0.590 & \textbf{0.570} & 0.580 & 0.602  &0.621 & 0.611 &-& -& -& -& - \\
{CNN-WARP} &    0.593 & 0.525 & 0.557 & 0.598 &  0.614 & 0.607 &-& -& -& -& - \\
{CNN-RNN} &     0.660 &0.556 &0.604 &0.692&\textbf{ 0.664} &0.678 &-& -& -& -& - \\
BCE    &    \textbf{0.804} & 0.547 & 0.651 & 0.815  &0.605 & {0.695} &0.3357& 0.0191& 30.6551& 0.0243& 0.6584\\
C2AE &    0.428 & 0.411 & 0.409 & 0.474 & 0.528 & 0.499 & 0.4560& 0.4809& 36.1900 &0.1416& 0.4512 \\
\hline
RETDM &  0.799 &0.555 & \textbf{0.655} &\textbf{0.819} & 0.611 & \textbf{0.700} &\textbf{0.3295} &\textbf{0.0189} &\textbf{30.3492}& \textbf{0.0219} & \textbf{0.6628}\\
\hline
\end{tabular}
\label{coco}
\end{table*}

\subsection{Experimental Setting}

To verify the effectiveness of RETDM, we compare it with the following related methods:
\begin{itemize}
\item LMMO-kNN \cite{Liu2018Metric,Liu2015Large}\footnote{The code is downloaded from the authors' homepage: https://sites.google.com/site/weiweiliuhomepage/.}: large margin multi-output metric learning with $k$ nearest neighbor constraints (LMMO-kNN) that is a shallow deep metric learning paradigm to incorporate the predefined loss functions to learn the embedding space based on $k$ nearest neighbor constraints.
\item MLSPL \cite{Li2018A}: MLSPL integrates a self-paced learning strategy to learn instances and labels from an easy-to-hard fashion, which is proposed recently and shows promising results for multi-label learning.
\item C2AT \cite{Yeh2017Learning}\footnote{The code is downloaded from https://github.com/yankeesrules/C2AE.}: Canonical Correlated AutoEncoder  uniquely integrated  deep canonical correlation analysis (DCCA) and autoencoder in a unified DNN model, which is recently proposed for multi-label learning.
\item CL \cite{Wen2016A}: CL is a powerful feature learning approach for face recognition that  simultaneously learns a center for deep features of each class and penalizes the distances between the deep features and their corresponding class centers. Actually, CL can be regarded as a deep distance metric learning method. Therefore, we compare with it in the experiment.
\item BCE: it treats each label as independent one, and trains one classifier for each label. In the experiment, we use VGG-16 with binary cross-entropy as the network architecture.
\end{itemize}

In our method, there are some parameters, such as $\alpha$, $\beta$, and the number of nearest neighbors $k$ that needed to be set in advance.
The parameters $\alpha$, $\beta$ in our method are chosen by cross validation. The number of the nearest neighbors $k$  is set to 10 throughout the experiments.

In order to sufficiently verify our method, we utilize extensive criteria mentioned in \cite{Wu2017A}.
We first evaluate the effectiveness of the proposed approaches with the following five criteria: hamming loss, ranking loss, one error, coverage, and average precision. These criteria are commonly used for evaluating multi-label learning algorithms \cite{Zhou2008Multi,Huang2013Fast}.
\begin{itemize}
\item Hamming loss: The hamming loss evaluates how many times an sample-label pair is misclassified, i.e., a wrong label is predicted.
The performance is perfect when the hamming loss is equal to zero; the smaller the value of the hamming loss, the better the performance of the model.
\item Ranking loss: It evaluates the average fraction of label pairs that are not ordered correctly for the sample. The performance is perfect when the ranking loss is equal to zeros; the smaller the value of the ranking loss, the better the performance of the model.
\item one error: It measures how many times the top-ranked label is not a correct label of the sample.
The performance is perfect when one error is equal to zero; the smaller the value of one error, the better the performance of the model.
\item coverage: It measures how far it is needed, on the average, to go down the list of labels in order to cover all the correct labels of the sample.
The smaller the value of coverage, the better the performance of the model.
\item average precision: It evaluates the average fraction of correct labels ranked above a particular threshold. The performance is perfect when the average precision is equal to one; the larger the value of the average precision, the better the performance of the model.
\end{itemize}

Besides the above five criteria, we also compute macro precision (denoted as ``C-P''), micro precision (denoted as ``O-P''),  macro recall (denoted as ``C-R''), micro recall (denoted as ``O-R''), macro F1-measure (denoted as ``C-F1''), and micro F1-measure (denoted as ``O-F1'').  ``C-P'' is evaluated by averaging per-class precisions, while ``O-P'' is an overall measure that counts true predictions for all images over all labels.
Similarly,  ``C-R'' and ``O-R'' can be also evaluated.
The F1 (``C-F1'' and ``O-F1'') score is the geometrical average of the precision and recall scores.

Note that since the above criteria measure the performance of the model from different aspects, it is difficult for one algorithm to outperform another on every one of these criteria. However, in our experiment, our method outperforms other state-of-the-arts in most of the criteria.

\subsection{Experimental Results}

\begin{figure*}[htb]
\centering
\subfigure[C-P]{\includegraphics[width=0.32\linewidth]{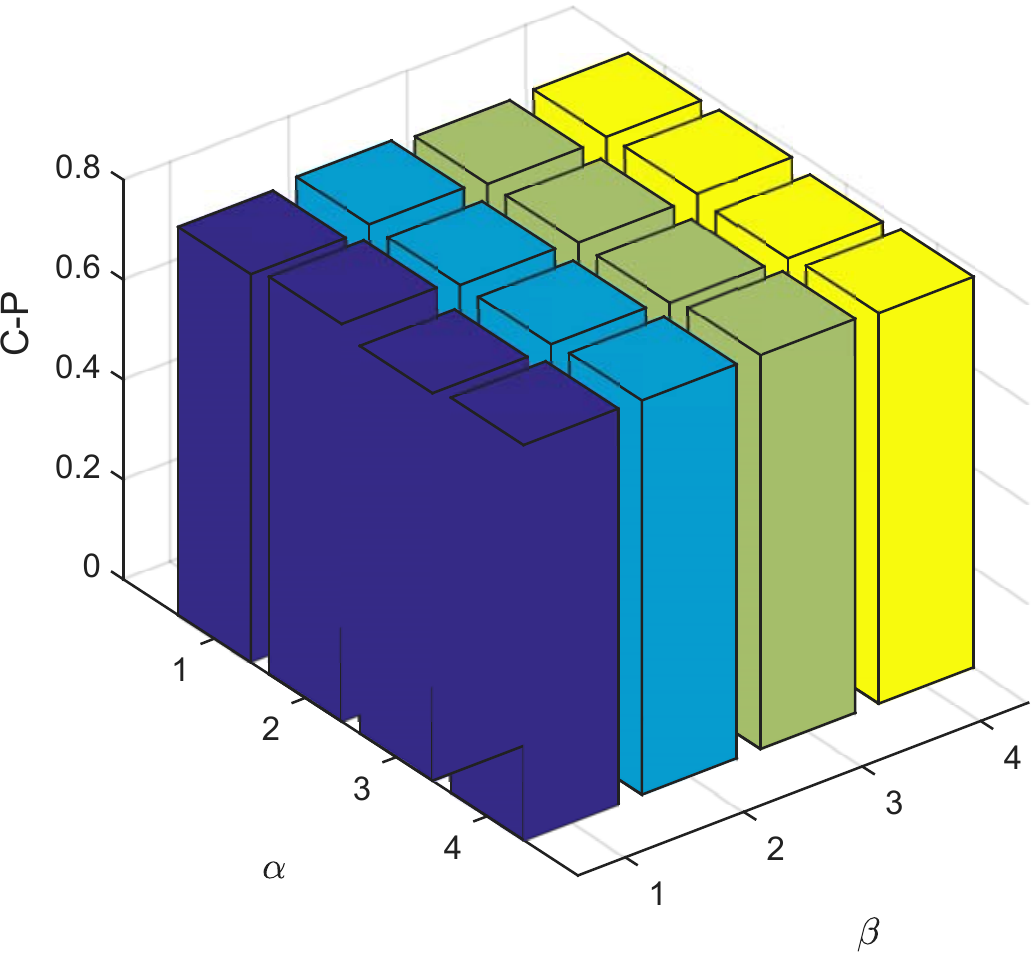}}
\subfigure[C-R]{\includegraphics[width=0.32\linewidth]{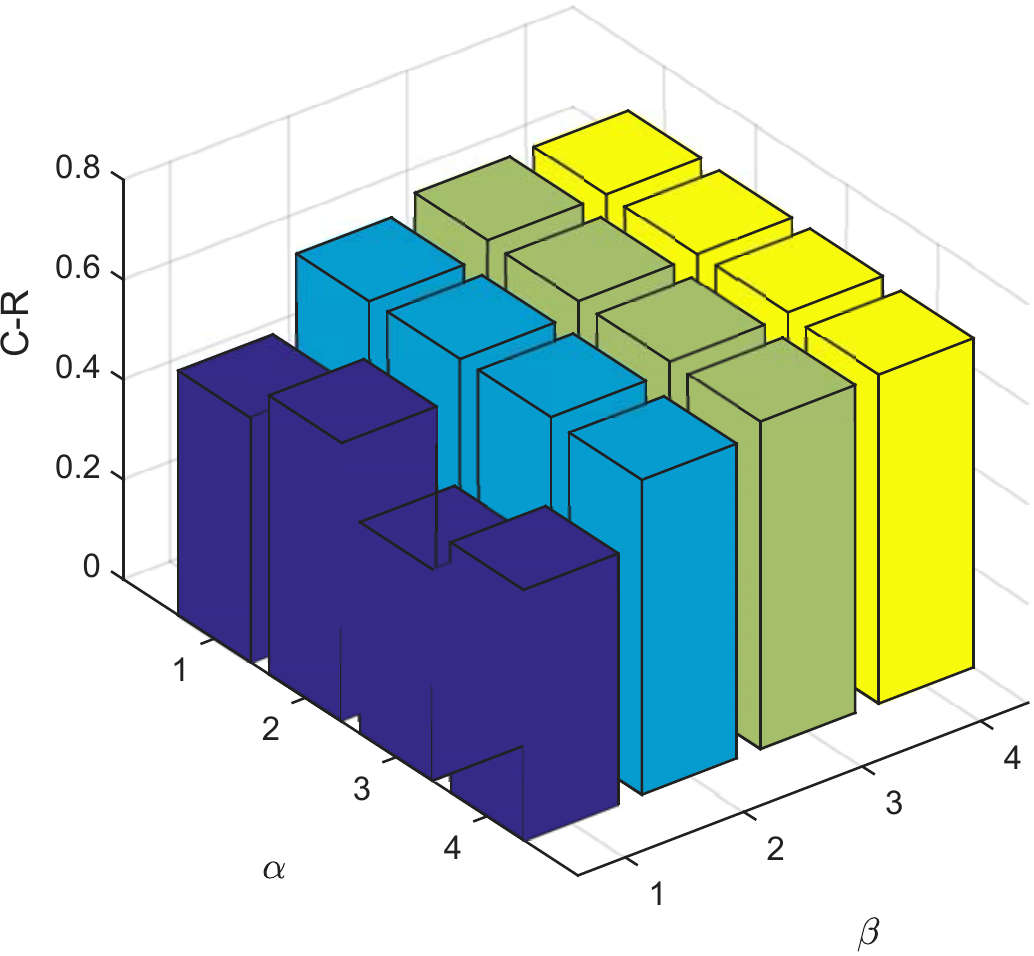}}
\subfigure[C-F1]{\includegraphics[width=0.32\linewidth]{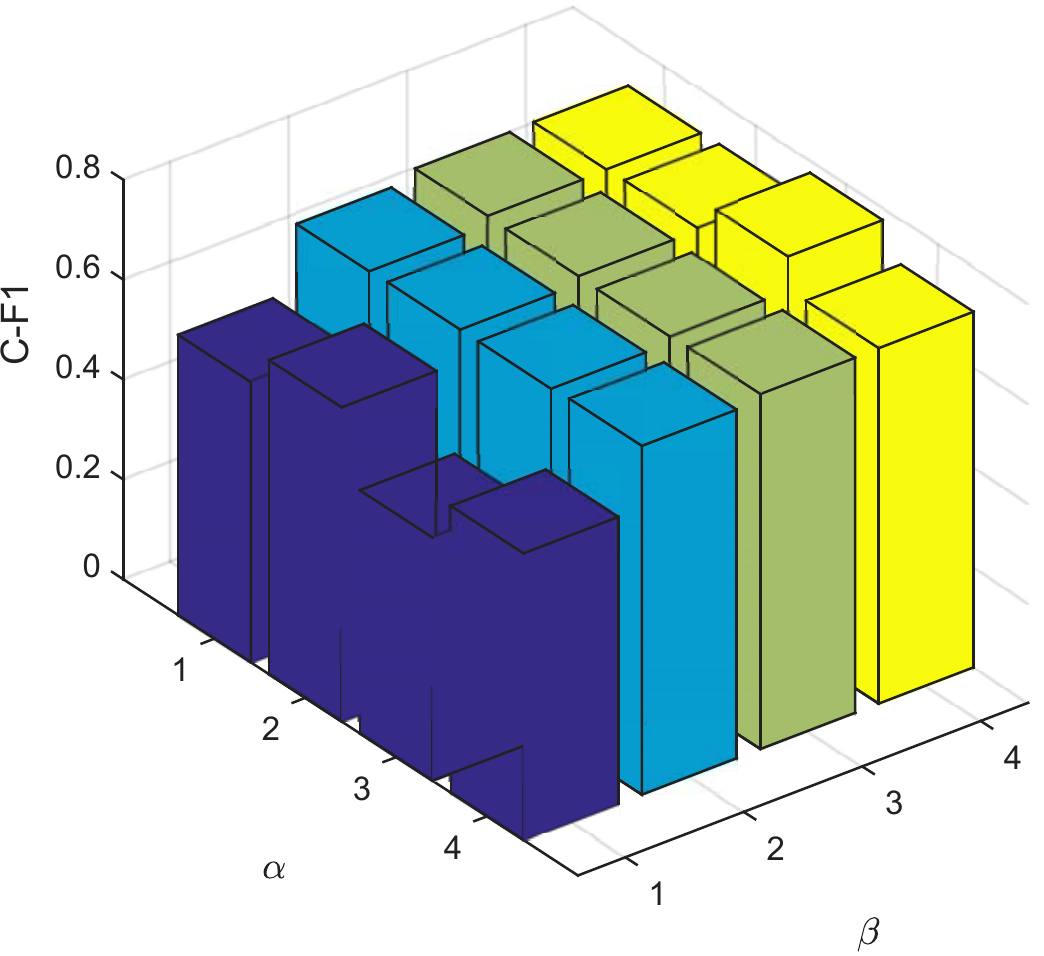}}
\subfigure[O-P]{\includegraphics[width=0.32\linewidth]{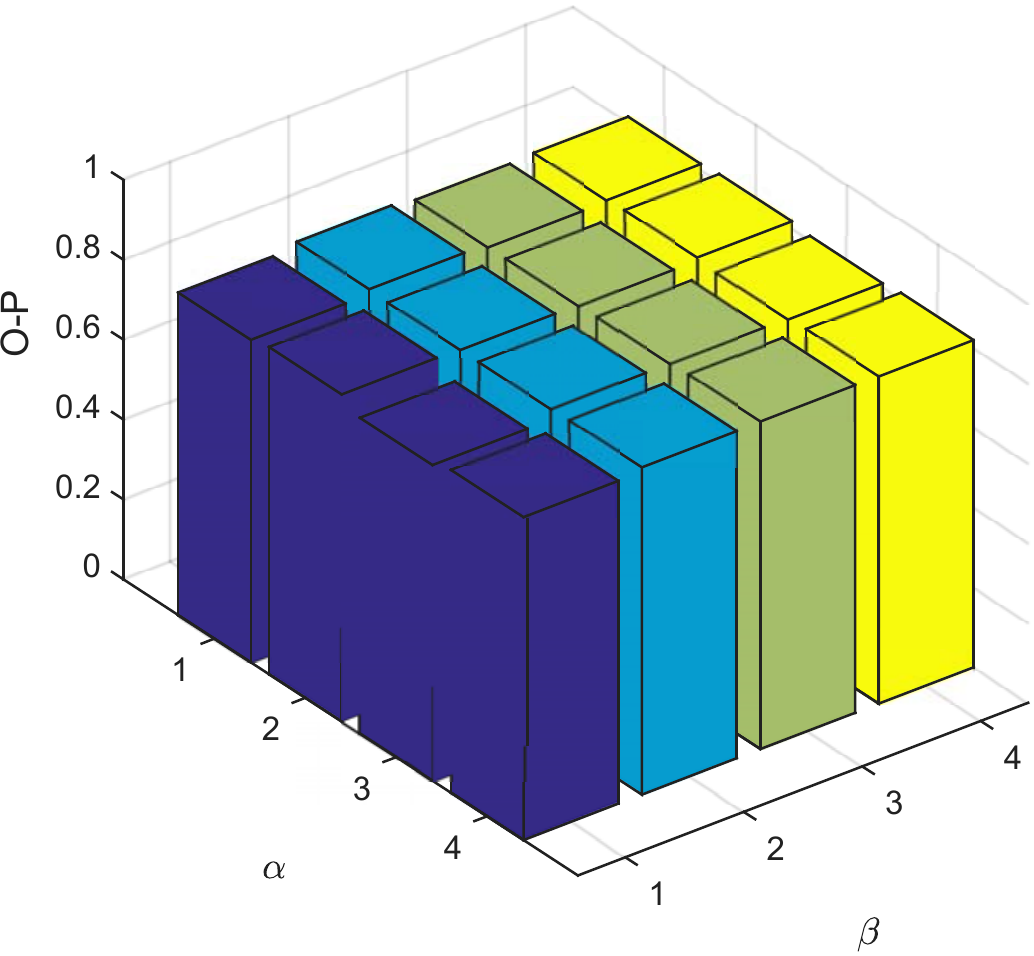}}
\subfigure[O-R]{\includegraphics[width=0.32\linewidth]{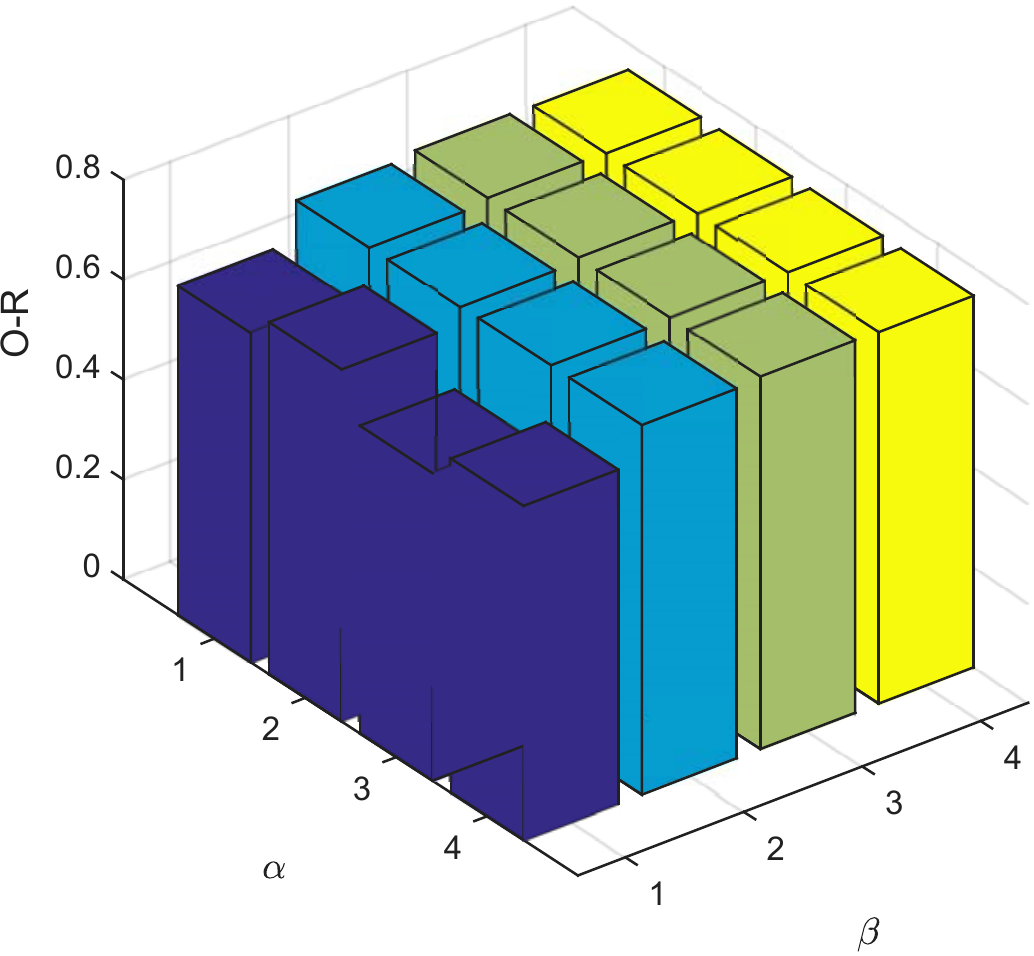}}
\subfigure[O-F1]{\includegraphics[width=0.32\linewidth]{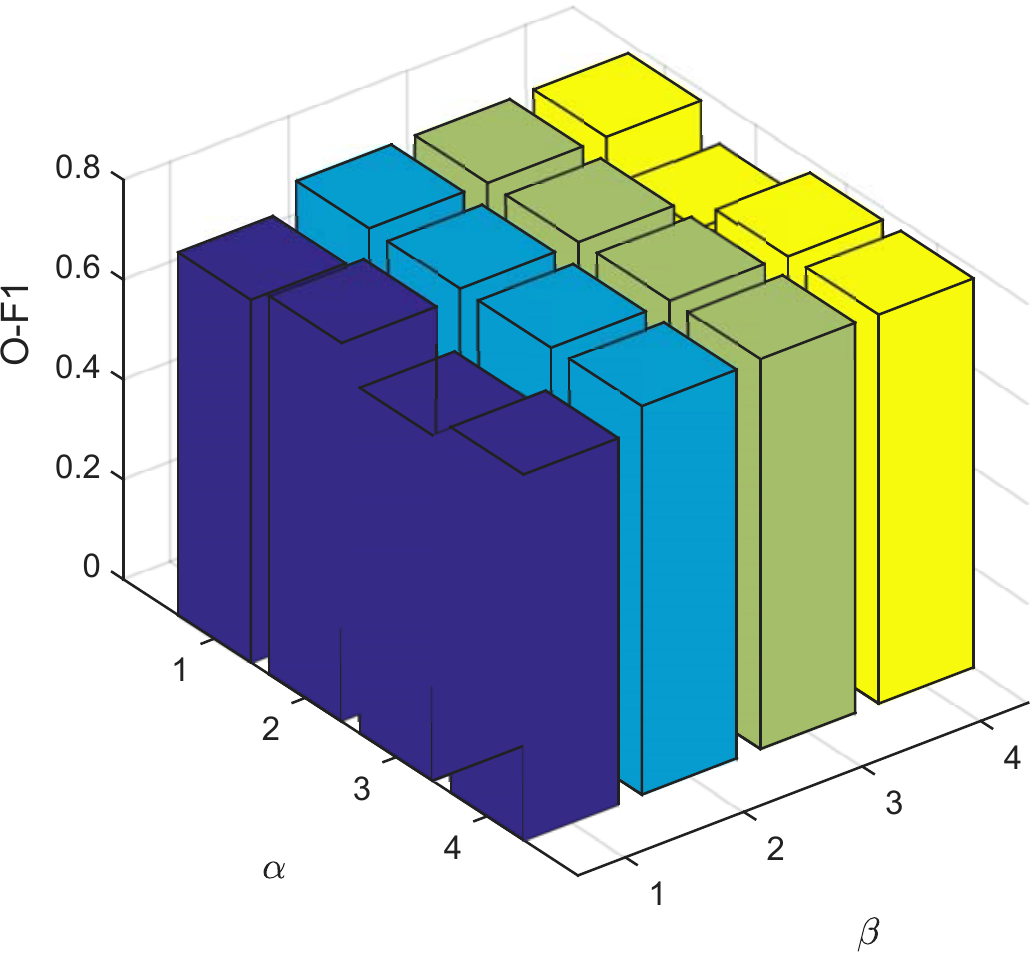}}
\caption{Sensitivity study of the parameters in terms of ``C-P'', ``C-R'', ``C-F1'', ``O-P'', ``O-R'', and ``O-F1'' on the mirflickr dataset.}
\label{sensitive}
\end{figure*}

\begin{table*}
\caption{Verify the effectiveness of two-way distance metric module on the MS-COCO dataset.}
\label{performance}
\centering
\begin{tabular}{c|c|c|c|c|c|c||c|c|c|c|c}
\hline
 {Method}  & C-P  & C-R  & C-F1  & O-P  & O-R  & O-F1 & hamming loss & ranking loss & coverage &  one error& average precision \\
\hline
RETDM (one-way) &  \textbf{0.806} & 0.545 & 0.651 &\textbf{ 0.828} & 0.601 & {0.696} & 0.3389& \textbf{0.0189}& 30.9184&\textbf{0.0215}& 0.6575 \\
\hline
RETDM &  0.799 &\textbf{0.555} & \textbf{0.655 }&0.819 & \textbf{0.611} & \textbf{0.700} &\textbf{0.3295} &\textbf{0.0189} &\textbf{30.3492}& 0.0219 & \textbf{0.6628}\\
\hline
\end{tabular}
\label{freq}
\end{table*}

We first test the general performance of our method RETDM on the three  image datasets.
Tables I-III summarize the results of different methods in terms of all the eleven evaluation criteria.
From these tables, we can see that our method RETDM significantly outperforms LMMO-kNN and MLSPL on the three datasets, which indicates that deep network model indeed has better performance than the shallow models in the scenario of multi-label image classification.
In the meantime, RETDM achieves better results than C2AE and BCE. This shows our method can better exploit the correlations of labels to improve the performance.
Finally, RETDM beats CL on all the three datasets, which demonstrates our deep metric learning model can learn more discriminative distance metric for multi-label image classification.
In addition, the performance of CL is quite unstable in terms of all criteria. It indicates that the deep metric learning for other applications can not be directly applied to multi-label classification.
Note that we do not run CL on the MS-COCO dataset, because of its prohibitively training cost.

In our paper, we propose a two-way distance metric learning module. In this section, we verify its effective. To do this, we add another experiment on the MS-COCO dataset. In the experiment, we only use one-way strategy to learn the metric, i.e., the loss function $\mathcal{J}_{metric}$ is equal to $\mathcal{J}_{1}$ in Eq. (\ref{loss1}). We name it ``RETDM (one-way)'' for short. The results are listed in Table IV.
From the Table, we can see that RETDM is better than RETDM (one-way) with most of the criteria. This illustrates our two-way strategy is good for the multi-label classification problem.

\begin{figure}
\centering
\subfigure[mirflickr]{\includegraphics[width=0.47\linewidth]{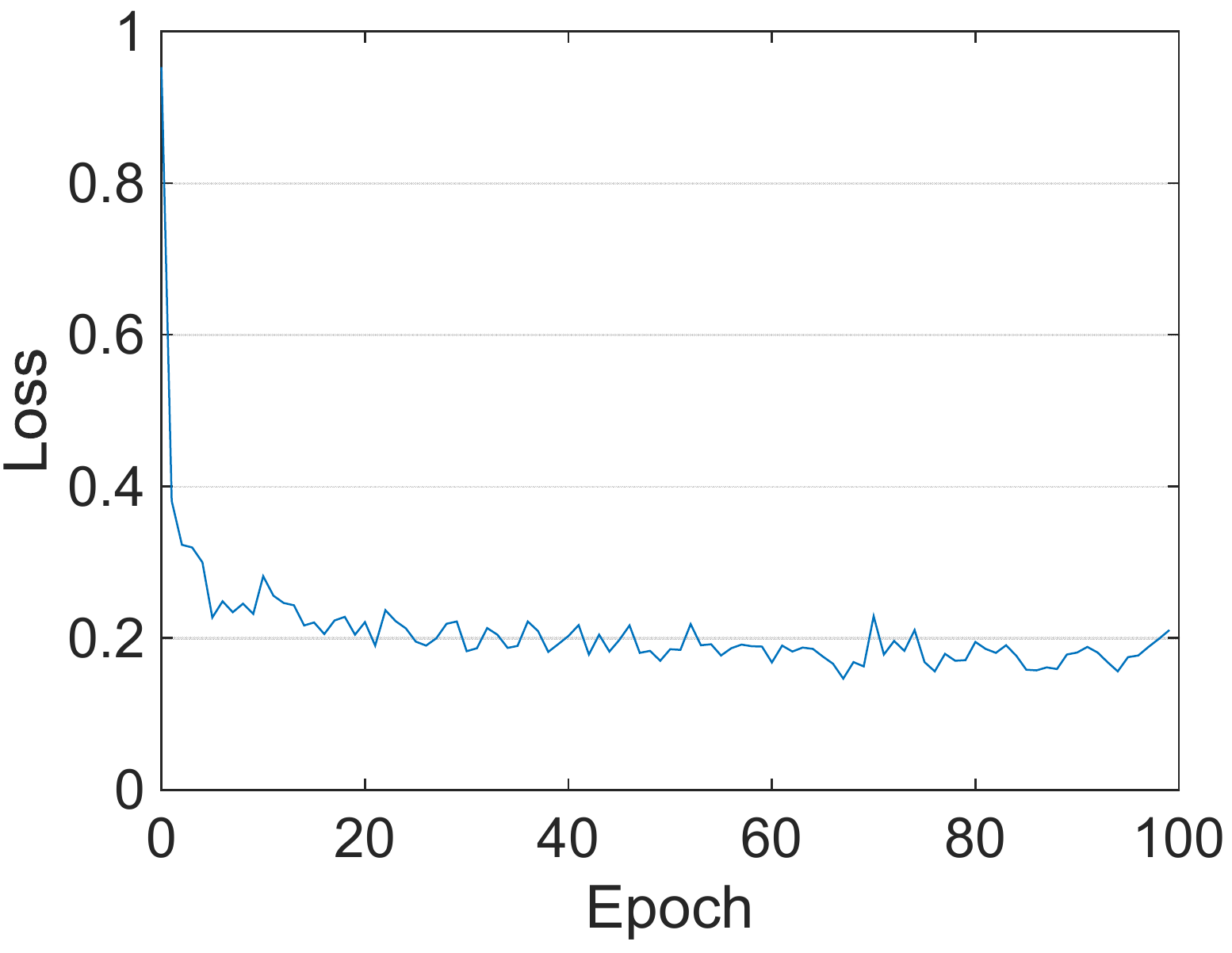}}
\subfigure[MS-COCO]{\includegraphics[width=0.5\linewidth]{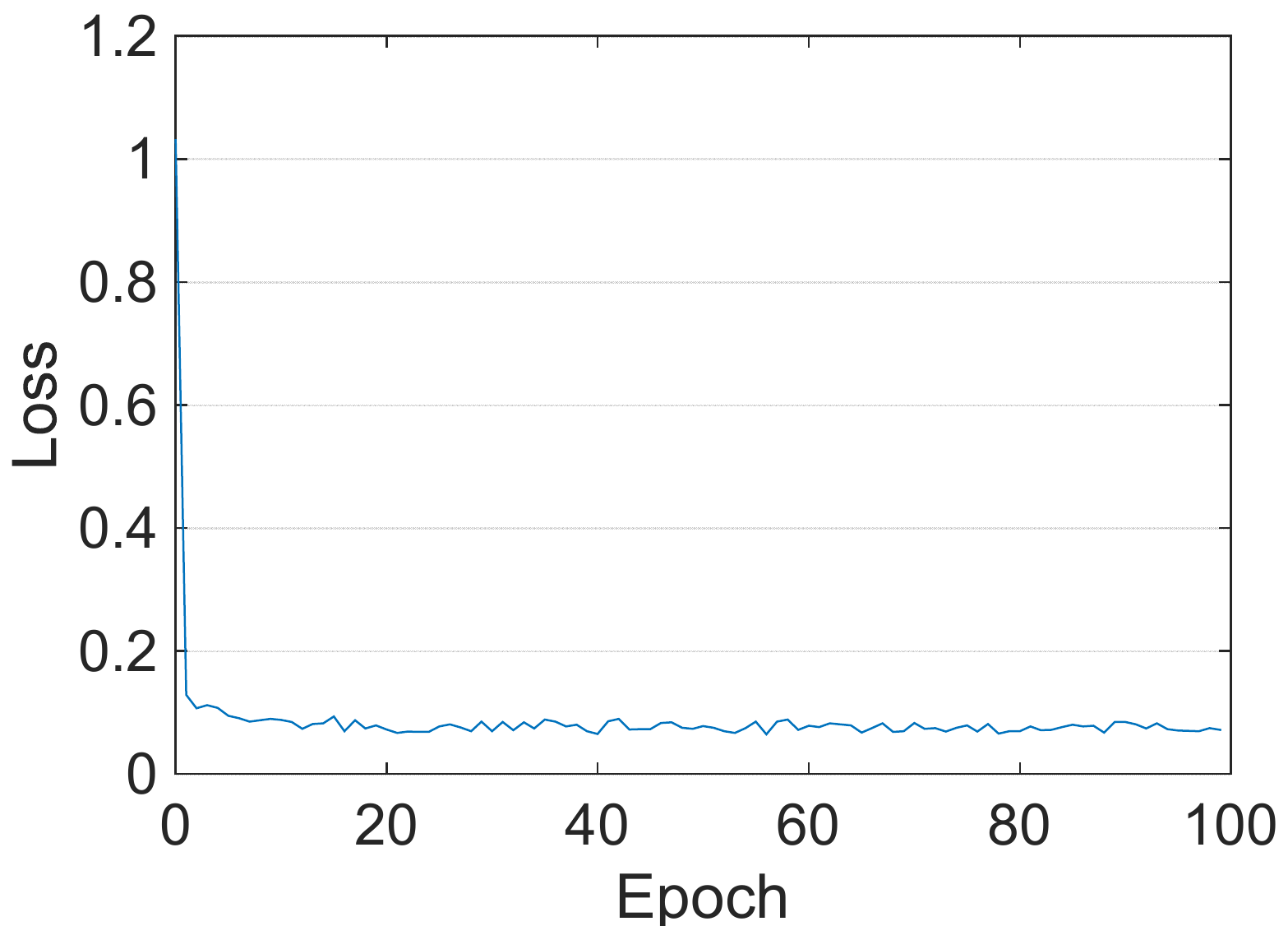}}
\caption{Convergence analysis on the mirflickr and MS-COCO datasets.
}
\label{convergence}
\end{figure}

We also study the sensitivity of parameters $\alpha$ and $\beta$ in our algorithm on the mirflickr dataset.
Fig. \ref{sensitive} shows the results. From Fig. \ref{sensitive}, our method is not sensitive to $\alpha$ and $\beta$ with wide ranges.
Fig. \ref{convergence} shows the convergence curve of our method on the mirflickr dataset.
As in Fig. \ref{convergence}, we can see RETDM has a good convergence rate.
It will converge after only about 20 epochs.

\section{Conclusion}
In this paper, we proposed a novel deep distance metric learning framework for multi-label image classification, named RETDM.  First, RETDM aimed to learn a latent space to embed images and labels via two independent deep neural networks respectively, so that the input dependency and output dependency can be well captured.
After that, a reconstruction regularized deep metric network was presented for mining the correlations between input and output and making the embedded space more discriminative as well.
Extensive evaluations on scene, mirflickr, and NUS-WIDE datasets showed that our proposed RETDM significantly outperformed the state-of-the-arts.

Several interesting directions can be followed up, which are not covered by our current work. For example, we can leverage input's nearest neighbors in our approach. RETDM learns a two-way distance metric learning based on target label's $k$ nearest neighbors, as well as $k$ input images with their labels being $k$ nearest neighbors of the target labels.
Symmetrically, input image's $k$ nearest neighbors and their corresponding target labels can be also involved into our method. A potential issue is that we need to find  input image's $k$ global nearest neighbors during each iteration, which would increase the burden of computation.

\bibliographystyle{IEEEtran}
\bibliography{aaai2018}

\end{document}